\documentclass[letterpaper, 10 pt, conference]{IEEEtran}
\IEEEoverridecommandlockouts

\usepackage{amsmath}
\usepackage{moreverb}
\usepackage{cite}
\usepackage{amsmath,amssymb,amsfonts}
\usepackage{algorithmic}
\usepackage{graphicx}
\usepackage{textcomp}
\usepackage{hyperref}
\usepackage{xcolor}
\usepackage{subfig}
\usepackage{multirow}
\usepackage{float}

\floatstyle{ruled}
\newfloat{pddl}{thp}{lop}
\floatname{pddl}{PDDL listing}

\def\BibTeX{{\rm B\kern-.05em{\sc i\kern-.025em b}\kern-.08em
    T\kern-.1667em\lower.7ex\hbox{E}\kern-.125emX}}

\title{PlanSys2: A Planning System Framework for ROS2}

\author{
    \IEEEauthorblockN{Francisco Mart\'in}
    \IEEEauthorblockA{\textit{Intelligent Robotics Lab} \\
    \textit{Rey Juan Carlos University}\\
    francisco.rico@urjc.es}
    
    \and

    \IEEEauthorblockN{Jonatan Ginés Clavero}
    \IEEEauthorblockA{\textit{Intelligent Robotics Lab} \\
    \textit{Rey Juan Carlos University}\\
    jonatan.gines@urjc.es}
    
    \and
    \IEEEauthorblockN{Vicente Matell\'an}
    \IEEEauthorblockA{\textit{Robotics Group} \\
    \textit{University of León}\\
    vicente.matellan@unileon.es}
    
    \and

    \IEEEauthorblockN{Francisco J. Rodríguez}
    \IEEEauthorblockA{\textit{Robotics Group} \\
    \textit{University of León}\\
    fjrodl@unileon.es}
}

\begin{document}

\maketitle

\begin{abstract}

Autonomous robots need to plan the tasks they carry out to fulfill their missions. The missions' increasing complexity does not let human designers anticipate all the possible situations, so traditional control systems based on state machines are not enough. This paper contains a description of the ROS2 Planning System (PlanSys2 in short), a framework for symbolic planning that incorporates novel approaches for execution on robots working in demanding environments. PlanSys2 aims to be the reference task planning framework in ROS2, the latest version of the {\em de facto} standard in robotics software development. Among its main features, it can be highlighted the optimized execution, based on Behavior Trees, of plans through a new actions auction protocol and its multi-robot planning capabilities. It already has a small but growing community of users and developers, and this document is a summary of the design and capabilities of this project.

\end{abstract}

\section{Introduction}

For decades, symbolic Planning has been a significant field within Artificial Intelligence, being applied since the first robots to plan the tasks they had to perform \cite{Munson1971RobotPE}\cite{Nagata73}. Since then, this has been a field of active research that has never lost its relevance over the years \cite{georgeff1987reactive}\cite{haigh1998interleaving}. Much of its popularity is thanks to the creation of planning languages such as STRIPS~\cite{Fikes1971STRIPSAN} or its successor PDDL~\cite{McDermott1998PDDLthePD}, which establish a common framework to integrate the planning algorithms produced by the scientific community. As planning languages become standardized, the existence of planning competitions~\cite{Vallati2015} where researchers can create benchmarks where they can measure and compare planning algorithms has been possible. Among the planning algorithms that come from this field, we can list POPF (Forward-Chaining Partial-Order Planning) \cite{ICAPS101421} and TFD (Temporal Fast Downward) \cite{Eyerich2012}, which are used in PlanSys2.

In robotics, planning systems for robots integrate knowledge management, planning, and execution plans in real and simulated robots. ROSPlan~\cite{ICAPS1510619} is the most used planning system in the last decade. It has had 28 contributors and more than a thousand commits throughout the last six years of development. It has fueled dozens of projects in as many robots, both land \cite{Miranda2018ARM}\cite{fmartin18}, air \cite{Nogueira18},  and marine \cite{Palomeras16}. Much of the success of ROSPlan is because it is the reference package for Planning in ROS, which is the \emph{de facto} standard in robot programming. ROS \cite{Quigley2009} has a community of thousands of programmers and users and has been adopted by hundreds of companies that use this Open Source framework as the base of their developments. 

In this work, we address the needs of new robots and applications, creating a planning system that can be widely used in many applications that require reliability, performance, and flexibility. Initially planned as a successor to ROSPlan, PlanSys2\footnote{http://intelligentroboticslab.gsyc.urjc.es/ros2\_planning\_system.github.io} far exceeds the applications and features of its predecessor. First of all, it is built on top of ROS2\footnote{https://design.ros2.org}, ROS's evolution towards a more modern, predictable, better-designed framework with many more features. ROS2 also aims to fulfill the requirements of the industry to be applied in critical and real-time systems. PlanSys2 takes advantage of the new features of ROS2, such as lifecycle nodes or multicast communications in real-time that DDS\footnote{https://www.omg.org/spec/DDS} provides.

PlanSys2 brings several novel contributions to planning systems. The plans generated by the planning algorithms are transformed into Behavior Trees~\cite{Colledanchise18} (a mathematical model of task execution that combines efficiency and flexibility) using a novel algorithm~\cite{fmartin21} for plan analysis and parallel execution of action flows. PlanSys2 also incorporates a novel algorithm for delivering actions to action performers (those components that execute an action) through auctions. This approach allows to have a multi-robot execution and to specialize action performers in handling certain specific elements. PlanSys2 also offers the ability to merge different PDDL domains to support modular applications. 



\section{Related Work}
\label{sec:related}


Automatic generation of plans has been used in Robotics for several decades~\cite{Nourbakhsh-1996-16261}. Shakey~\cite{Nilsson84} (1966-1972) was the first robot that used Symbolic Planning. This robot was used for testing STRIPS \cite{Fikes71}, one of the first languages for AI Planning. Rhino robot approach \cite{Burgard88} generated an action plan to guide visitors through the Bonn museum, and Dervish~\cite{Nourbakhsh_Powers_Birchfield_1995} automatically calculated his action plan at the AAAI 1994 Robot Competition, winning the Office Delivery event. However, the mechanisms associated to these systems had computational limitations. The difficulty of integrating planning and action in robots working in complex and dynamic environments led researchers such as Arkin \cite{Arkin89} and Brooks \cite{Brooks91} to seek new techniques for generating robotic behaviors far from symbolic planning based approaches.

This trend is changing in recent years due to the increase in the computing power of robots, and the problems of subsymbolic systems to cope with explainability requirements, or to deal with the symbolic interaction with humans. Cognitive architectures like CORTEX \cite{luis-pablo-cortex} embed symbolic planners to bring plans to reality. Major problem of these architectures is that their planning system are not directly usable in another architecture, and that it is very difficult also to change its planners.

The emergence of standarized robot programming frameworks has favored the emergence of a planning system. The most widespread is ROSPlan~\cite{ICAPS1510619}, which has been the reference in ROS. ROSPlan is the inspiration for PlanSys2, taking its capabilities far beyond. By running on ROS2, PlanSys2 already has a large number of features related to security and predictability. ROSPlan is only capable of controlling one robot, and, in our experience, using it in various robotics research and competitions~\cite{app10176067}, it is not as efficient as we wished.

SKiROS~\cite{skiros} is another ROS framework that uses Planning. Besides, it has an ontology-based reasoning engine, which is an idea that we want to incorporate into PlanSys2 soon. Like ROSPlan, it has not been designed for multirobot and is primarily intended towards executing simple manipulation tasks.

\section{PlanSys2 Design}
\label{sec:design}

PlanSys2 has been designed to be robust and to be able to scale to adapt to new functionalities. We intend that it becomes a useful de facto standard for researchers and industry. Researchers in classical planning will have an experimentation framework in which they can integrate their algorithms and test the execution of their plans in simulator applications or real robots. For researchers in Robotics, they will expand the components to support aspects such as logical reasoning, ontologies, knowledge representation, or different control paradigms. Robot companies will be able to count on a reliable and efficient framework to implement complex applications in the real world. The design has also taken into account aspects necessary for real applications such as multi-robot coordination, cybersecurity, quality of service, or safety.

\subsection{Reliable, Secure and Predictable}

PlanSys2 is built on top of a real-time capable meta-operating system, ROS2, to address operational safety standards and determinism. ROS2 is the new version of the ROS framework, which aspires to be a standard in Robotics, as its predecessor is de facto. Its predecessor's main difference is that it uses the Data Distribution Service (DDS) communication standard used in critical systems such as spacecraft, airplanes, hyperloop, or next-generation automobiles. Including DDS as the glue of its components, ROS2 is endowed with new capabilities that already make it suitable for industrial-grade mobile robotics with high requirements in real-time, safety, and security. ROS2 exposes the DDS configuration in terms of strict quality of service in real-time communications and its security features.

ROS2 also introduces the concept of Managed Nodes (also known as LifeCycle Nodes). These nodes have a clear life cycle based on the states and transitions from their creation to their destruction. These states are observable, and their transitions are triggered internally and externally. The startup process of the nodes that are part of an application is deterministic. Each state and transitions have a clear responsibility that includes memory allocation, configuration, communications management, central processing, or error handling, making the software that uses them predictable. All PlanSys2 components and the implementation of the actions that the robot must perform are LifeCycle Nodes.

\subsection{Modular and Extensible}

\begin{figure}[tb]
  \centering
  \includegraphics[width=\linewidth]{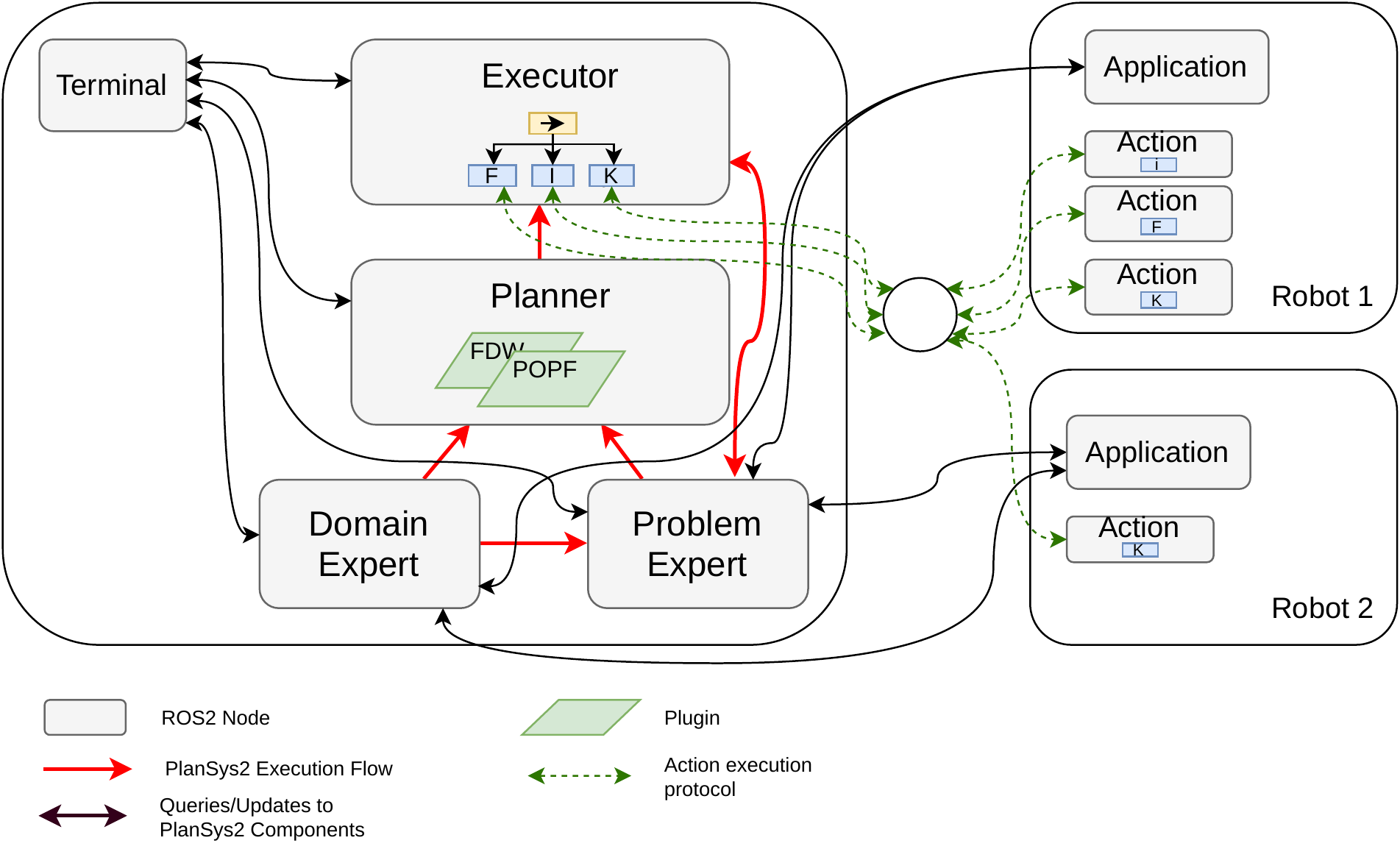}
  \caption{PlanSys2 architecture.}
  \label{fig:plansys2}
\end{figure}

For PlanSys2 to become widely used, it must adapt to the needs of users and developers. For this reason, we have made a modular design that we show in figure \ref{fig:plansys2}. Each component clearly defines its interfaces and can be replaced by another with different implementations or capabilities.

The central component of this design is the \emph{Planner node}, which calls the \emph{plan solver} that contains the planning algorithm. It is designed to be able to use different plan solvers through the use of plugins. The default solver plan is POPF, although TFD is also available. PlanSys2 can be extended to new plan solver by a plugin that contains how to call it and parse the generated plan. Each time the Planner Node is asked to generate a plan, it relies on the \emph{Domain Expert} node and the \emph{Problem Expert} node to provide the domain, and the problem, respectively, in PDDL format.

The \emph{Domain Expert} reads the PDDL domain file and stores it internally in memory. It is capable of accepting several different domains that we can mix to get a single one. This capacity allows for modular applications where each package contributes its part of the domain and the implementation of the actions it requires. This component's function is to provide the domain to the planner when requested and to receive requests for information about the domain, such as whether a type exists or whether a predicate is valid with particular instances.

The \emph{Problem Expert} contains the problem's status to be solved, including instances, predicates, functions, and goals, always validated with the Domain Expert node. In general, all the dynamic knowledge of the application. This component composes the PDDL problem when requested by the Planner node. How we store this knowledge (memory, database) or any reasoning about it has no limitation as long as it can meet this request.

The \emph{Executor Node} is in charge of making plans generated by the Planner node come true. When PlanSys2 is used in a specific application, it receives requests from it to execute the plan that arises with the knowledge (including goals) that exists at that moment in the Problem Expert Node. To do this, it asks the Planner node for a plan and, if it exists, executes it. We turn the plan into a Behavior Tree that carries it out, checking requirements and applying effects at runtime. To execute the actions, we have designed an auction protocol that selects the most appropriate node that implements the action to be performed.

The applications that use PlanSys2 contain the nodes that implement the actions and the PDDL model that implements them. Any application also includes a controller node that accesses the Problem Expert's knowledge to consult and establish instances, predicates, and goals. This controller is also the one that requests the Executor node to execute or cancel plans.

\subsection{Support for Multi-robot and Specialized Action Performers}
\label{sec:multi}
PlanSys2 is not limited to generating plans for a single robot. We have designed a system that is capable of executing plans on different robots and devices. To make this feature a reality, we have relied on the ability of ROS2 nodes to communicate with practically zero-configuration (since DDS uses Multicast and in ROS2, all nodes of the same network can communicate with each other) and a new action auction protocol. There can be multiple nodes in a network that implement the same PDDL action. Each of them can specialize in executing actions with specific parameters. 

For example, each robot in a group of robots would have a node that implements the \small\texttt{move }\normalsize action, which would make the robot navigate between areas of its environment. Each robot would specialize in executing only those actions whose first parameter matches its identifier, like in, for example, \small\texttt{(move rb1 area1 area2)}\normalsize. When the Executor must execute the move action, he would start an auction to see who performs it, to which only those nodes that implement the action, which has specialized in the parameters in which it is specialized, would bid. 

This protocol is not limited to robot selection; it can also be specialized in certain elements. Consider, for example, the possibilities that this offers for manipulation when we can find the action that best implements actions such as \small\texttt{(pick rb1 apple)}\normalsize or \small\texttt{(pick rb1 cup)}\normalsize.

\subsection{Efficient and Explainable}

The plans generated by the plan solver have room to be executed efficiently. In PlanSys2, we consider the plan solvers as black boxes, so the executor's optimization is carried out on the generated plan. We analyze the plans to identify the execution flows of a plan. This mechanism allows actions to be executed in parallel as soon as the preceding actions are finished.

Each update of the status of any component of PlanSys2 publishes comprehensible updates to which any visualization or monitoring application can subscribe: knowledge updates, generated plans, the status of the execution of a plan, feedback on each of the action execution, and the messages of the auction protocol. The goal is to make PlanSys2 transparent, explainable, and debuggable.

\section{PlanSys2 Implementation}
\label{sec:implementation}

In this section, we will focus on the details and implementation of the newest PlanSys2 contributions. These main contributions to state-of-the-art planning systems focus on executing actions in two aspects: The execution of plans such as Behavior Trees and the action execution protocol. We also will include in this section some relevant implementation contributions.

\subsection{Efficient execution of Plans as Behavior Trees}

The Behavior Trees is a mathematical model of performance control that has become popular in recent years for controlling robots due to its flexibility and reactivity. It is a tree made up of nodes, with leaves being those that do not have a child node. The nodes contain control structures that affect their children: sequential execution, parallel execution, condition checking, fallbacks, or retries, among others. The leaves of the tree contain perceptual processing or control generation. These leaves can also make requests for information or activation to other components of the system. The basic operation that can be performed on a node is the tick, which the control nodes can cascade to call their children. This operation can have a return value of failure, success, or running, which indicates that a node's mission has not yet been completed. A tree is executed by ticking its root as long as the return value is running.

BTs are ideal for executing plans since they allow to express any plan as execution in sequence or in parallel of the leaves (this is not entirely exact, as we will see below, but it is established here for simplicity) that represent the execution of the actions of the plan.

\begin{figure}[ht!]
  \centering
  \includegraphics[width=0.8\linewidth]{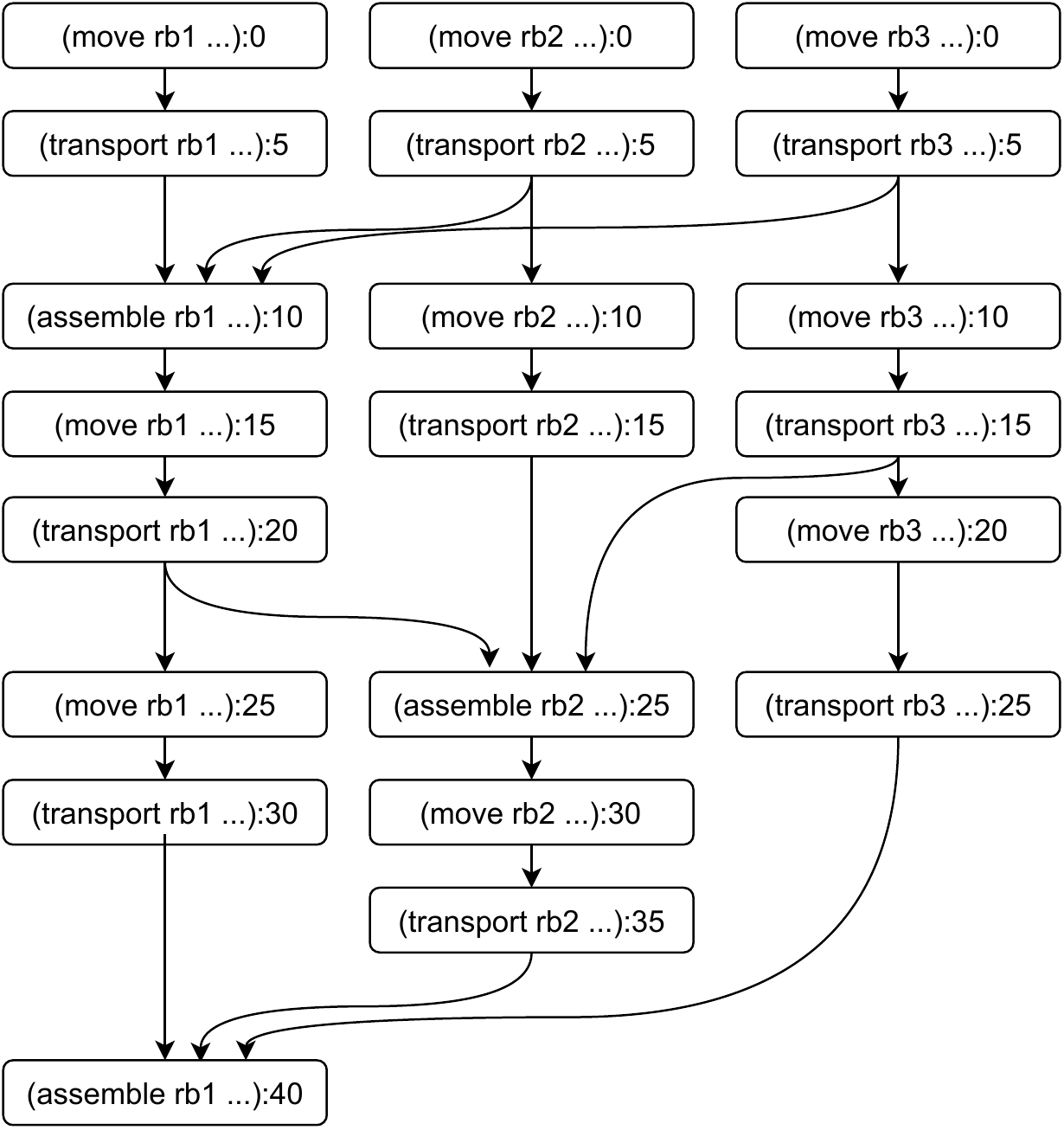}
  \caption{Plan graph generated from Listing \ref{pddl:sciroc_plan}}
  \label{fig:planflow}
\end{figure}

Let's consider a PDDL plan like the one shown in Listing \ref{pddl:sciroc_plan}. In this plan, three robots collaborate to assemble three cars. The first step is to analyze the plan and build an execution graph. This graph (shown in Figure \ref{fig:planflow} represents the dependency between actions, from which we can extract the execution flows. 

\begin{pddl}[h!]
\scriptsize  
\begin{verbatimtab}
0	(move rb1 assembly_zone body_car_zone)
0	(move rb2 assembly_zone steerwheel_zone)
0	(move rb3 assembly_zone wheels_zone)
5.001	(transport rb1 bc_1 body_car_zone assembly_zone)
5.001	(transport rb2 stwhl_1 steerwheel_zone assembly_zone)
5.001	(transport rb3 whl_1 wheels_zone assembly_zone)
10.002	(assemble rb1 assembly_zone whl_1 bc_1 stwhl_1 car_1)
10.002	(move rb2 assembly_zone body_car_zone)
10.002	(move rb3 assembly_zone steerwheel_zone)
15.003	(move rb1 assembly_zone wheels_zone)
15.003	(transport rb2 bc_2 body_car_zone assembly_zone)
15.003	(transport rb3 stwhl_2 steerwheel_zone assembly_zone)
20.004	(transport rb1 whl_2 wheels_zone assembly_zone)
20.004	(move rb3 assembly_zone body_car_zone)
25.005	(assemble rb2 assembly_zone whl_2 bc_2 stwhl_2 car_2)
25.005	(move rb1 assembly_zone steerwheel_zone)
25.005	(transport rb3 bc_3 body_car_zone assembly_zone)
30.006	(move rb2 assembly_zone wheels_zone)
30.006	(transport rb1 stwhl_3 steerwheel_zone assembly_zone)
35.007	(transport rb2 whl_3 wheels_zone assembly_zone)
40.008	(assemble rb1 assembly_zone whl_3 bc_3 stwhl_3 car_3)
\end{verbatimtab}
\caption{\label{pddl:sciroc_plan}The plan generated to assemble cars with three robots.}
\normalsize
\end{pddl}

\begin{figure*}[tb]
  \centering
  \includegraphics[width=0.9\linewidth]{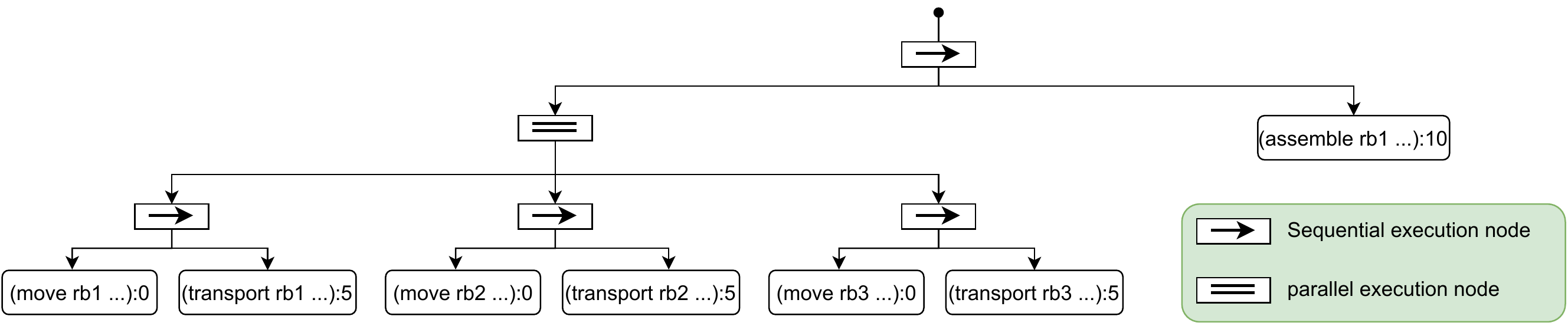}
  \caption{Plan graph generated from Listing \ref{pddl:sciroc_plan}}
  \label{fig:bt_plan}
\end{figure*}

All nodes in the graph that do not depend on one another are considered the source flows. In this example, the move actions scheduled for at time 0 mark the start of three flows that can be considered parallel. Throughout this graph, we can see flows that converge at a node (the assemble nodes) and others that diverge (the transport nodes). An action cannot be started until the actions that produce its incoming flows have been executed. We have developed an algorithm \cite{fmartin21}\cite{martin2021optimized} that guarantees that we can convert any graph into a BT that optimizes its execution in terms of parallelization. Figure \ref{fig:bt_plan} shows the portion of BT created from the graph in Figure \ref{fig:planflow} up to the Assemble action at 10. This BT contains nodes whose children must be executed in sequence and others whose children must be executed in parallel.

\begin{figure}[ht]
  \centering
  \includegraphics[width=0.85\linewidth]{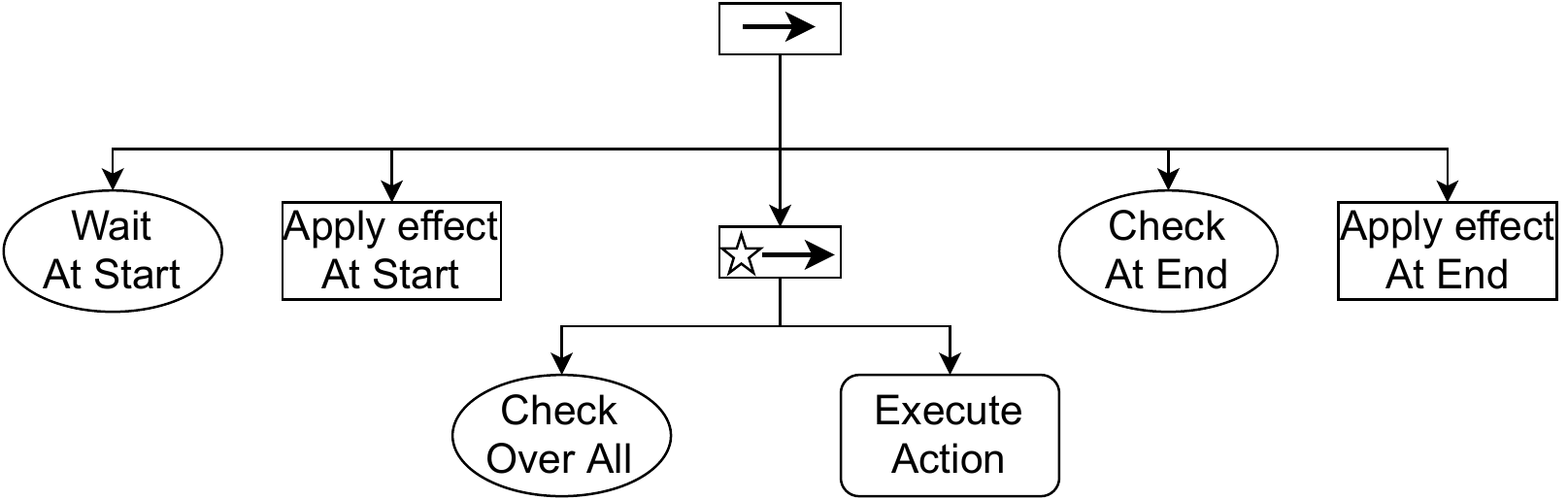}
  \caption{Expansion of each action unit.}
  \label{fig:bt5}
\end{figure}

Previously we had clarified that the leaves of this BT were not actions. In the last step of creating this BT, each action is replaced with a subtree that controls the execution of an action plan. In PDDL, a \emph{durative action}~\cite{fox2003pddl2} has requirements at the beginning, during, and end of an action. Besides, some effects apply at the beginning of the action execution and the end. The subtree shown in Figure~\ref{fig:bt5} does these functionalities:
\begin{itemize}
    \item A node that that returns running until the \small\texttt{at start }\normalsize  requirements are met.
    \item A node that applies the effects \small\texttt{at start}\normalsize, updating the knowledge in the Problem Expert with the predicates that the action indicates as effects at the beginning of the action.
    \item A \emph{star sequence}\footnote{https://www.behaviortree.dev/sequencenode/} node with a node that checks the \small\texttt{over all }\normalsize requirements and a node that commands the execution of an action.
    \item A node that checks if the \small\texttt{at end }\normalsize requirements are met.
    \item A node applies the effects \small\texttt{at end}\normalsize.
\end{itemize}

This structure can also be easily extended to include more nodes for debugging, quality assurance, or any functionality added to each action. One of the reasons to choose BTs is the flexibility to have these extensions.

Any node that returns \emph{failure} is transmitted to the BT root, causing the plan to fail. This situation may be because the requirements are not met or because the action fails.

\subsection{Action Execution Auction protocol}

PlanSys2 contributes an action execution protocol called "action auction"~\cite{lagoudakis2004simple}. We can find the nodes responsible for executing an action through this protocol and entrust them with this duty. Some of these nodes can be configured to only serve requests with certain parameters. For example, a node that implements the \small\texttt{move }\normalsize action can serve only those requests whose first parameter is \small\texttt{rb1}\normalsize.

\begin{figure}[tb]
  \centering
  \includegraphics[width=\linewidth]{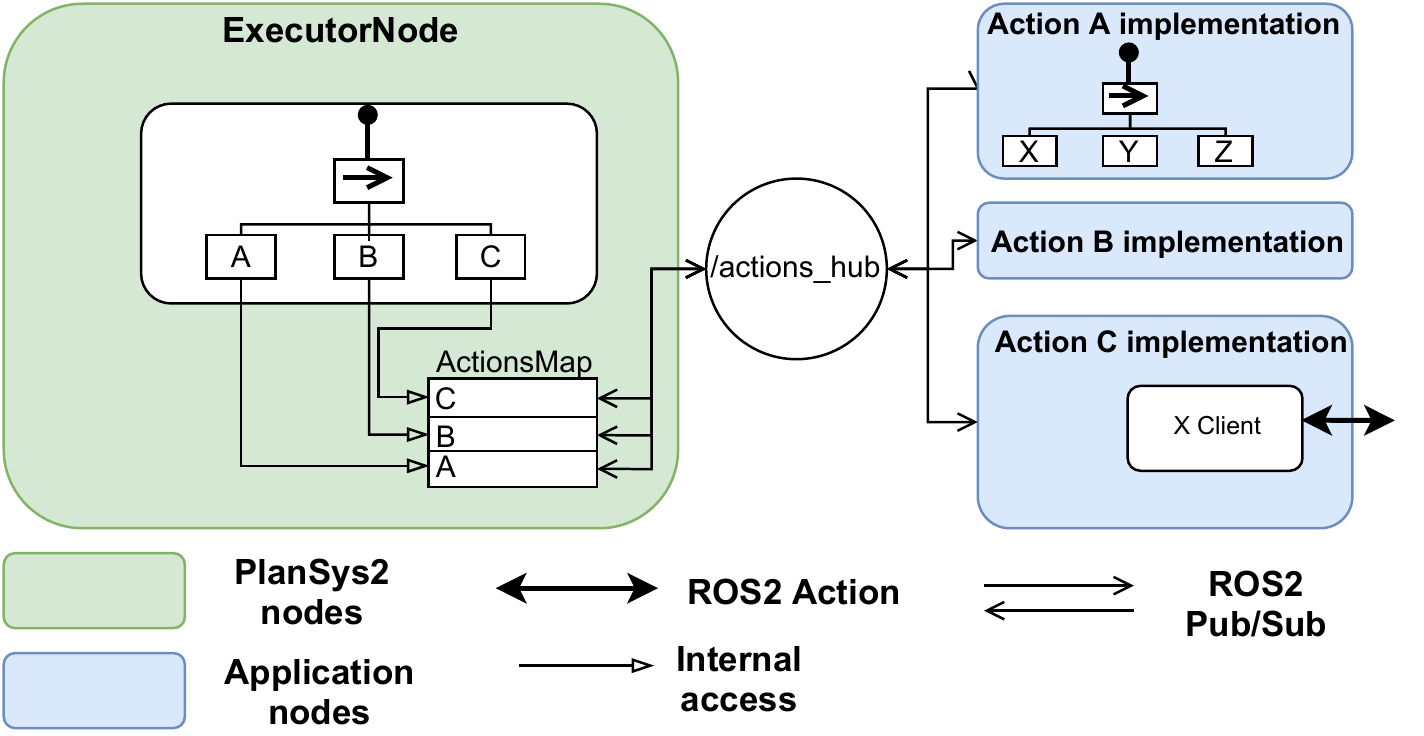}
  \caption{Plan Execution Flow.}
  \label{fig:plansys2_execution}
\end{figure}

In the previous section, we introduced a node in Figure \ref{fig:bt5} that executed an action. When this node is ticked for the first time, a new entry containing an \emph{ActionPerformerClient} is added to the \emph{ActionsMap} table (Figure \ref{fig:plansys2_execution}) to control the execution of this action. This object implements this protocol in the Executor side. The nodes that implement the actions are \emph{ActionPerformer}, which implements this protocol on this side of the communication. The channel through which this protocol is carried out is \footnote{action\_hub}\normalsize, to which both sides of the protocol publish and listen. As we said before, these nodes are in the LifeCycle node in an idle state, waiting to be assigned a task. Each node implements a single PDDL domain action.

\begin{figure*}[tb]
  \centering
  \includegraphics[width=0.58\linewidth]{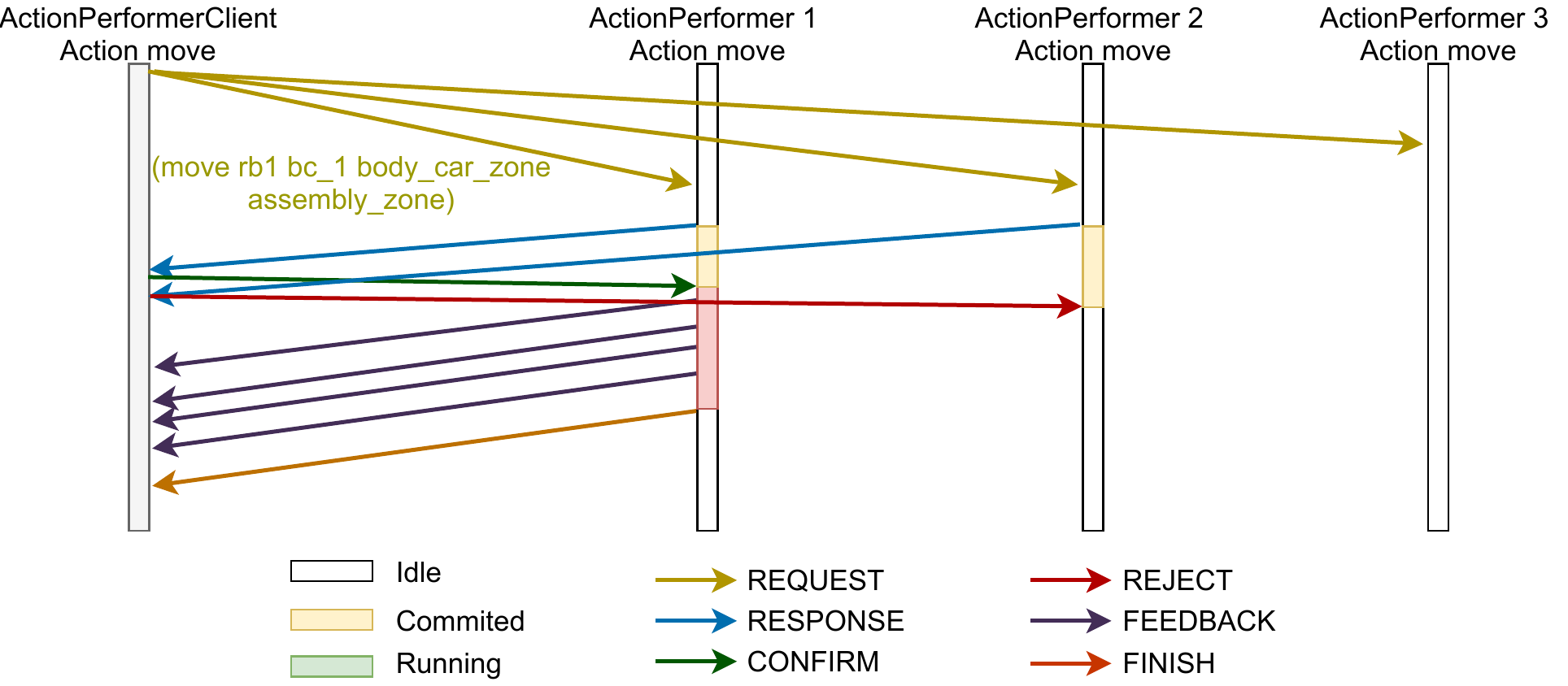}
  \caption{Plan Execution Flow.}
  \label{fig:protocol}
\end{figure*}

This protocol uses the topic \small\texttt{/action\_hub,}\normalsize in which both the \emph{ActionPerformerClient} and the \emph{ActionPerformer} publish and receive. This protocol works as follows:
\begin{itemize}
    \item When an \emph{ActionPerformerClient} is created, it sends a REQUEST message, specifying the action and its parameters. It will perform multiple retries if it does not respond in a reasonable amount of time (1 second, for example).
    \item Those nodes that are idle that implement this action and consider that they can carry it out return a RESPONSE message identify themselves. The ActionPerfomer 3 in the situation illustrated in Figure \ref{fig:protocol} does not attend the request because it is configured to only attend certain requests, as commented in \ref{sec:multi}.
    \item Until they are satisfied with a CONFIRM message or a rejected with a REJECT, they are committed to not participate in parallel in other auctions.
    \item The node that has been confirmed begins to execute. This involves changing the state, as the LifeCycle node, to active.
    \item Periodically, this node sends feedback messages, identifying itself and identifying the action to which it refers. Let's not forget that \small\texttt{/action\_hub }\normalsize may be publishing results of other actions or even new auctions.
    \item The execution ends with a FINISH message, which includes the action execution's final result.
    \item When a plan is aborted because some node of the BT returns a failure, or because its cancellation is requested, all the \emph{ActionPerformerClient} send a CANCEL message to abort the execution of the actions in progress.
\end{itemize}

\subsection{Action performers}

PlanSys2 offers facilities for the implementation of actions. As can be seen on the right of Figure~\ref{fig:plansys2_execution}, if the action uses the ROS2 actions to request the operation of some subsystem (as in the case of Nav2~\cite{macenski2020marathon2} or MoveIt~\cite{8793898}), PlanSys2 provides mechanisms that hides its development complexity.

A generic action has also been implemented that is configured with a Behavior Tree to be executed. Only the leaves of this Behavior Tree have to be implemented. This mechanism allows us to reuse these leaves in different actions and easily modify those actions.

\subsection{Utilities}

The main tool in PlanSys2 is the Terminal. This application is a shell that allows us to manage and monitor the status of PlanSys2. Among its features are:
\begin{itemize}
    \item Visualize the PDDL domain and the knowledge of the Problem Expert: predicates, instances, functions, and goals.
    \item View details about the predicates and actions of the PDDL domain.
    \item Set or delete instances, predicates, functions, and goals.
    \item Visualize and execute plans.
    \item Monitor the execution of a plan.
    \item Check the status of the nodes that implement the actions.
\end{itemize}

Being a shell and not requiring a graphical interface, it can be used from any terminal/shell, have several shells in operation, do it through ssh, or even in systems without a visual environment.

\section{Limitations}
\label{sec:limitations}

The first limitation of PlanSys2 is that PDDL support is not complete. For now, it supports PDDL 2.1, although it would be desirable to extend its capabilities to the latest version of PDDL. Also, for now, it only uses two plan solvers: POPF and TFD.
We hope that as PlanSys2 becomes popular and proves useful, the planning community can contribute with more PDDL solvers.

Another limitation is that the state's consistency in which the system remains when an action is aborted is not resolved. This problem is not only with PlanSys2 but with any planning system. When the initial effects have been applied, but not the final effects, the state may not be consistent. Still, there are strategies from the application controller to deal with this limitation by adding the needed predicates based on previous knowledge or the perception of the system.

Another limitation is that we do not take advantage of planners that offer an initial plan, and then improved plans can be generated when the previous plan is already running. We are working to make a smooth transition from executing one plan to another.

\section{Experiment and Analysis}
\label{sec:experiments}

The research question we want to answer in this experimentation: \emph{Is PlanSys2 robust and efficient enough to be applied with reliability in groups of real robots?}

To demonstrate the capabilities and robustness of PlanSys2, we have designed two experiments. The first experiment will be a simulation of how PlanSys2 executes plans on multiple robots and how this impacts the time to complete plans. The second is a robustness test on a real robot to show the system's reliability in a long-term experiment.

The metrics that we consider relevant to validate PlanSys2 and answer the research question stated are:

\begin{itemize}
    \item \textbf{Execution time, and the number of plans and actions executed}: The amount of time that the system has been running uninterruptedly generating plans and executing actions. This metric indicates that the robustness of the system is robust and can be used in long-term operations.
    \item \textbf{Efficiency}: This metric refers to the percentage of time spent running actions in PlanSys2 for the total execution time. From the execution time of a plan, efficiency will be calculated how long it is not dedicated to executing actions or planning. This measurement will indicate the Efficiency and reactivity of PlanSys2. When parallelizing actions, this value can be greater than 100.
    \item \textbf{Failures and re-planning}. These two metrics are opposite. While the failures indicate how many times the system has not been able to continue its execution, either due to system errors or the inability to execute actions, the re-planning indicates that the system can detect unfulfilled requirements at runtime and continue executing.
\end{itemize}

\begin{figure}[tb]
  \centering
  \includegraphics[width=0.7\linewidth]{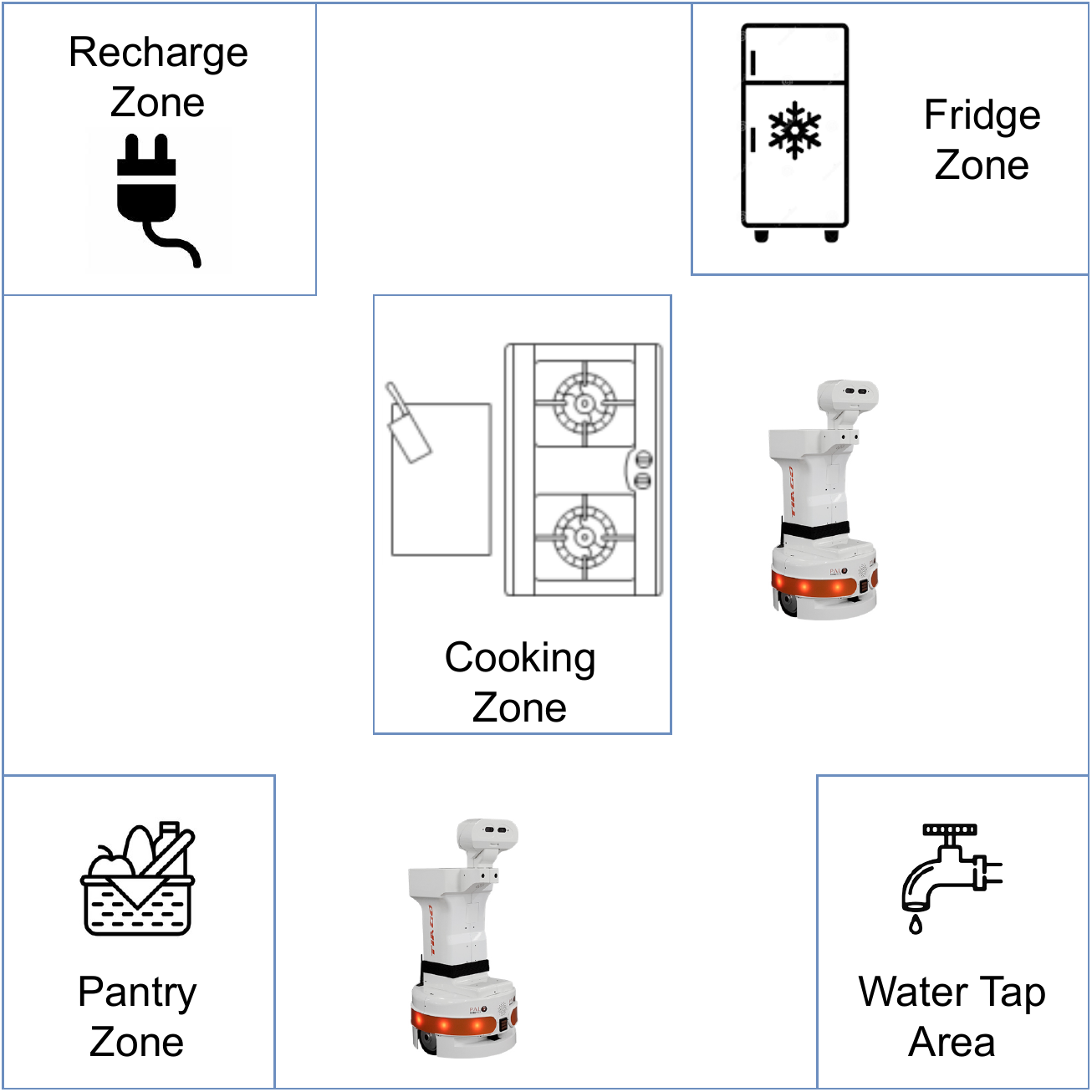}
  \caption{Plan execution flow.}
  \label{fig:exp-setup}
\end{figure}

The shared scenario for both experiments is shown in Figure \ref{fig:exp-setup}. A robot must cook three different types of dishes: cake, spaghetti, and omelet. The kitchen is in the center of the testing stage, while the ingredients are in 3 of the four corners. The robot must transport the ingredients to the cooking area and prepare the dish. Also, from time to time, the robot alerts for low battery. When this happens, the only action it can take is to move, and it must go to the recharging area to refill its battery and thus continue cooking. The system will generate requests composed of two random dishes. When the robot cooks the two dishes, two new requests are generated again. The actions are simulated, and their duration is approximately fixed: Move lasts 3.8 seconds, transport lasts 8.8 seconds (includes manipulation and movement) and cooking a dish for 21 seconds.

\subsection{Simulated multi-robot experiment}
\label{sec:multi-robot}

\begin{figure}[tb]
  \centering
  \includegraphics[width=0.75\linewidth]{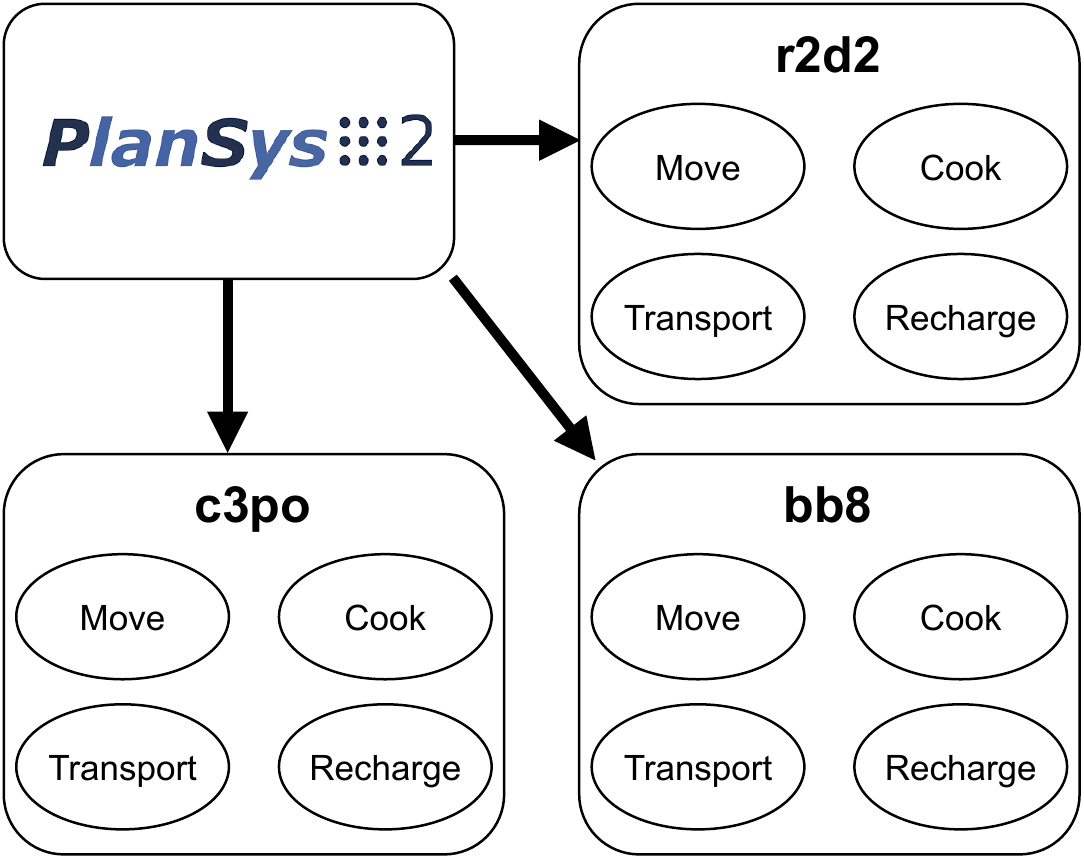}
  \caption{Software components for the multi-robot experiments. There are three robots (r2d2, c3po, bb8), each one with running implementations of all the actions.}
  \label{fig:exp1-sw}
\end{figure}

In this experiment, we want to show the control of a multi-robot system by PlanSys2. As shown in Figure \ref{fig:exp1-sw}, each of the robots executes an implementation of each of the four actions specified in the domain. All the actions have as their first parameter the robot that must perform them, so the nodes that execute the implementations of the actions are configured to attend only the actions for the robot in which it should be executed, following the principles outlined in section \ref{sec:multi}. PlanSys2 can be run on any computer on the network, including one of the robots.

The actions are simulated, and their duration is approximately fixed: Move lasts 3.8 seconds, transport lasts 8.8 seconds (includes manipulation and movement) and cooking a dish for 21 seconds.

Table \ref{tab:exp1} contains the experiment results carried out for one, two, and three robots. The total duration of each of them is around an hour and a half. In that hour and a half, a single robot has completed 76 plans, executing 1012 actions. It has not failed at any time. There have been nine replanning due to the robot running out of battery. 

The efficiency, which is 95\%, is a relevant metric, especially with one robot. This means that only 5\% of the time is dedicated to coordinate the start and stop executing the actions. We can conclude that PlanSys2 is very reactive because the elapsed time between executing consecutive actions is very low.

Increasing the number of robots allows many more plans to be executed. With plans of around 12-15 actions, the performance rises considerably for each added robot that cooperates to cook the dishes.

\begin{table}[ht]
\caption{\label{tab:exp1}Result of the experiment in simulation.}
\centering
\begin{tabular}{|c|c|c|c|}
\hline      & \textbf{1 robot} & \textbf{2 robots} & \textbf{3 robots} \\\hline 
Total Time & 8270 secs     &    8119 secs    &  8200 secs    \\\hline 
Plans & 76     &    101    &  135     \\\hline 
Actions & 1012    &    1550    &  2094     \\\hline 
Efficiency & 95.15 \%     &    137.54 \%    &  182.11 \%     \\\hline 
Fails & 0     &    0    &  0     \\\hline 
Replans & 9  &    22    &  31    \\\hline 
Dishes coocked & 154&    203    &  272     \\\hline 
\end{tabular}
\end{table}

This experiment's main conclusion is that the system has been in total seven and a half hours working without any failure. At that time, PlanSys2 carried out 312 plans.

\subsection{Real robot experiment}

\begin{figure}[tb]
  \centering
  \includegraphics[width=0.70\linewidth]{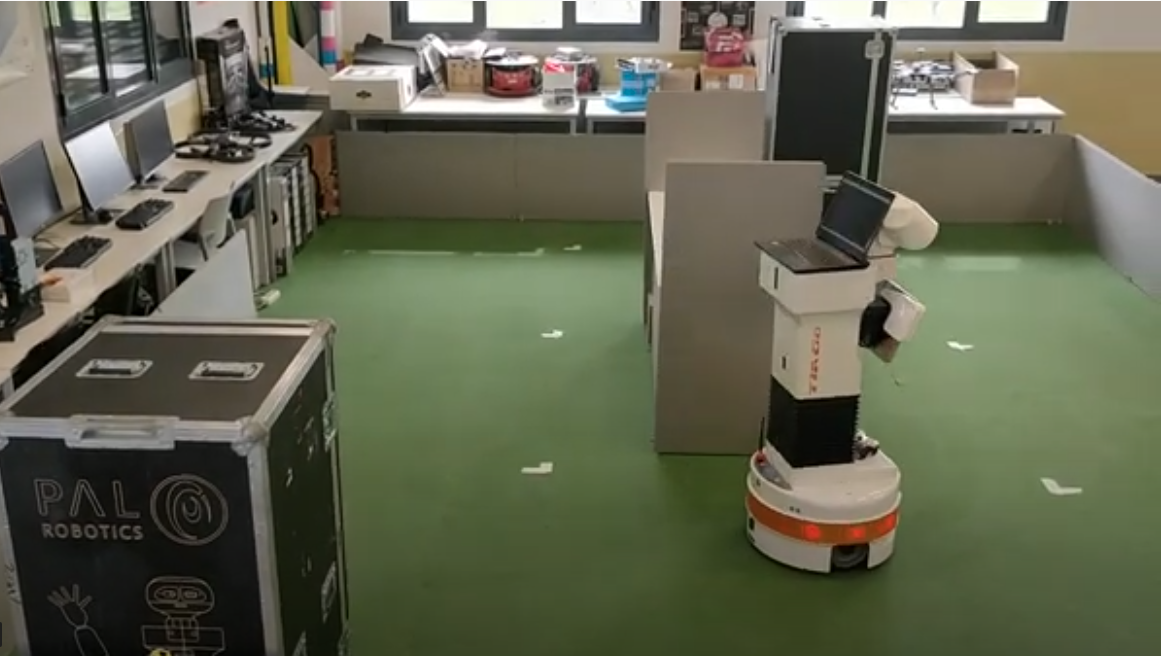}
  \caption{Long-time experiment in the real robot~\cite{exp_video}.}
  \label{fig:exp-long}
\end{figure}

We have set up a stage for the cooking experiment, as shown in figure \ref{fig:exp-long}. It is a stage with dimensions of 8x6 meters. We have placed the same arrangement of elements of figure \ref{fig:exp-setup}. We have used the TIAGo, a professional service robot. TIAGo runs ROS internally, so we have run ROS2, Nav2~\cite{macenski2020marathon2}, and PlanSys2 on a laptop above it. We use bridges to pass the messages between ROS and ROS2.

The robot does not have arms, so we have simulated only the part of the tasks that require them. Now that the robot must move, the tasks that involve movement will have a more realistic duration than in the previous experiment: Move takes 10-15 seconds, Transport 16-20 seconds, and Cook 21 seconds. As shown in Table \ref{tab:exp2}, the experiment lasted two hours, in which the robot carried out 32 plans, with no system failure and 9 re-plans. These replans are for simulated battery failures for the robot to head to charge. There were no failures in navigating from one point to another. The robot completed 1012 actions and a total of 72 dishes. As the actions have a duration closer to reality, the efficiency increases. Less than 3\% of the total time has been dedicated to managing PlanSys2 actions and plans.

\begin{table}[ht]
\caption{\label{tab:exp2}Result of the experiment with the real robot.}
\centering
\begin{tabular}{|c|c|}
\hline 
Total Time & 7216 secs \\\hline 
Plans & 32        \\\hline 
Actions & 432       \\\hline 
Efficiency & 97.19 \%      \\\hline 
Fails & 0       \\\hline 
Replans & 12    \\\hline 
Dishes coocked & 72    \\\hline 
\end{tabular}
\end{table}

This experiment's main conclusion is the validation of PlanSys2 in a real robot that has made dozens of plans without failure, so we have demonstrated its robustness.

The instructions to reproduce the experiment, both in real robot, or simulated are in this repository~\cite{exp_repo}.

\section{Conclusion and Future Works}
\label{sec:conclusions}

PlanSys2 has been described in this work, a new planning system for ROS2 that aspires to be a reference, as ROSPlan was in ROS. PlanSys2 takes advantage of the new features that ROS2 provides: A fast, secure, and reliable real-time communications framework and software design mechanisms that provide the predictability required in critical systems. Among the new features incorporated into PlanSys2, the ability to execute plans in several robots stands out, or to be able to select which software component is responsible for carrying out an action based on the parameters. Besides, we have incorporated a new plan execution engine based on Behavior Trees, which provides us with great flexibility and reactivity. Finally, we have described a protocol for executing actions to materialize multirobot and specialized execution. All of them with a system that provides enough information in real-time so that the system is traceable and explainable.

To validate its robustness and efficiency, we have conducted two experiments. The first experiment shows multi-robot execution. The second experiment shows the operation of PlanSys2 in a real professional robot. 10 hours of uninterrupted operation without any failure guarantee that PlanSys2 is ready to real and critical production systems.

There is still a lot to do in PlanSys2. A small development community has already been created around PlanSys2, which introduces improvements to adapt the systems to their own needs, such as debugging tools or increasing PDDL support. We hope to get the Planning scientific community's attention so that PlanSys2 encourages the Planning and Robotics communities come together.

\section{Acknowledges}
I want to thank the RoboCup@Home community for being the motivation of this development, the members of the Gentlebots team, and especially the PlanSys2 contributors, many of them introducing PlanSys2 in their companies or organizations: Fabrice Larribe (Safran Electronics \& Defense, France), Josh Zapf and Alexander Xydes (Space and Naval Warfare Systems Center Pacific, USA), and Stephen Balakirsky (Georgia Tech, USA). A warm thanks to Jacob Perron and Scott K. Logan (Open Source Robotics Foundation) for their assistance creating packages for ROS2 Eloquent and Foxy.

\bibliographystyle{IEEEtran}  
\bibliography{plansys2}  

\end{document}